\documentclass{article}
\usepackage{microtype}
\usepackage{graphicx}
\usepackage{subfigure}
\usepackage{booktabs} 
\usepackage{hyperref}
\usepackage{amsmath,graphicx,mlspconf}
\usepackage{amssymb}
\usepackage{mathtools}
\usepackage{amsthm}
\usepackage[capitalize,noabbrev]{cleveref}
\usepackage{caption}
\usepackage[lined,boxed,commentsnumbered]{algorithm2e}

\usepackage{dsfont}
\usepackage{float}
\usepackage{xfrac}
\usepackage{xcolor}
\usepackage{psfrag}
\usepackage{amsthm}
\usepackage{subfigure}
\usepackage{enumerate}
\usepackage{import}
\usepackage{epsfig}
\usepackage{amssymb}
\usepackage{pgfplots}
\usepackage{graphicx}
\usepackage{booktabs}
\usepackage{algorithm2e,setspace}
\usepackage{caption}
\usepackage{subcaption}
\usepackage{hyperref}

\DeclareMathAlphabet{\bm}{OML}{cmr}{bx}{it}
\DeclareMathAlphabet{\mathsf}{OT1}{cmss}{m}{n}
\DeclareMathAlphabet{\bs}{T1}{cmss}{bx}{sl}
\DeclareMathAlphabet{\ms}{T1}{cmss}{m}{sl}
\DeclareMathAlphabet{\mathpzc}{OML}{zplm}{m}{it}

\newcommand{\bg}[1]{\boldsymbol #1} 

\newcommand{\bb}{\mathbb}

\newcommand{\mc}{\mathcal}

\usepackage[textsize=tiny]{todonotes}

\title{Neural-ANOVA: Analytical Model Decomposition using Automatic Integration}

\name{Steffen Limmer, Steffen Udluft, Clemens Otte\thanks{This work was supported by the German BMBF under grant 03XP0389A and BMWK under grant 01IS24087A.}}
\address{Siemens AG, Foundational Technologies, Munich, Germany}

\begin{document}

\maketitle

\begin{abstract}


The analysis of variance (ANOVA) decomposition offers a systematic method to understand the interaction effects that contribute to a specific decision output. In this paper we introduce \emph{Neural-ANOVA}, an approach to decompose neural networks into the sum of lower-order models using the functional ANOVA decomposition. Our approach formulates a learning problem, which enables fast analytical evaluation of integrals over subspaces that appear in the calculation of the ANOVA decomposition. Finally, we conduct numerical experiments to provide insights into the approximation properties compared to other regression approaches from the literature.

\end{abstract}

\section{Introduction}

The functional analysis of variance (ANOVA) decomposition is an effective method to separate interaction effects between input variables and output variable with applications in industrial domains such as modeling of batteries \cite{adachi2023bayesian} and fluid flows \cite{yang2012adaptive}. A primary challenge in computing the ANOVA decomposition is the calculation of higher-dimensional integrals over subspaces of the input domain which is typically addressed by restricting the approximation space to analytically integrable functions such as random-forests \cite{hutter2014} or Fourier functions \cite{Potts2021}.

In this study, we introduce a novel method for calculating the ANOVA decomposition using standard neural networks in order to obtain a full decomposition with numerically accurate orthogonal components. We refer to these models as \emph{Neural-ANOVA} models. Our key contributions are as follows:

\begin{enumerate}
\item We introduce a novel learning formulation that enables fast analytical evaluation of integrals over subspaces appearing in the ANOVA decomposition of neural networks.
\item Through evaluations on test functions and datasets, we provide deeper insights into the expressive power and noise robustness of neural architectures under mixed differentiation and show that Neural-ANOVA compares favorably to standard regression approaches such as generalized additive models (GAMs).
\end{enumerate}

\begin{figure}
    \hspace{-0.5cm}
    \includegraphics[width=1.0\linewidth]{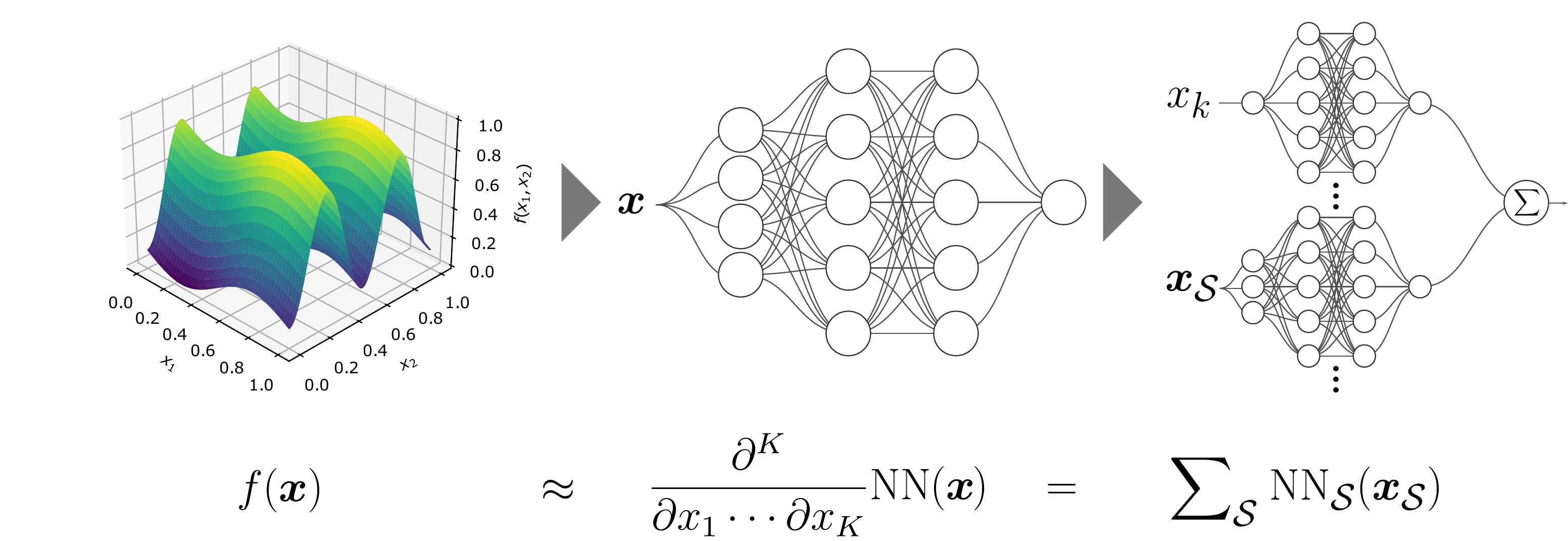}
\caption{Neural-ANOVA decomposition. The original data is approximated by the mixed partial derivative of a neural network (NN). An analytical ANOVA decomposition is obtained by decomposing the trained NN into lower-dimensional subnetworks $\text{NN}_\mc{S}(\bm{x}_\mc{S})$. These subnetworks are derived through closed-form evaluation of integrals over  subspaces.} 
    \label{fig:anova-teaser}
\end{figure}

\section{Related Work}

\subsection{Generalized and Neural Additive Models}
Generalized Additive Models (GAMs)~\cite{Hastie2017} are a powerful and versatile approach for machine learning problems that incorporate non-linear relationships between features and the target variable through a sum of univariate functions

\begin{equation}\label{eq:gam}
f(\bm{x}) = f_0 + \sum\nolimits_{k=1}^{K}f_k(x_k),
\end{equation}
or a sum of both uni- and bivariate functions
\begin{equation}
    f(\bm{x}) = f_0 + \sum\nolimits_{k=1}^{K}f_k(x_k) + \sum\nolimits_{k=1}^{K}\sum_{l<k}f_{kl}(x_k, x_l).
\end{equation}

A traditional method to train GAMs is \emph{backfitting} \cite{breiman1985estimating}, which iteratively updates the components of the model by sequentially refitting them.
More recently, \emph{Neural Additive Models} (NAMs) \cite{agarwal2021neural} using shape functions parametrized as neural networks and trained using stochastic gradient methods have been proposed.

\subsection{Functional ANOVA Decomposition}
The functional ANOVA decomposition \cite{hoeffding1948central,sobol2001global,hooker2004discovering} is a statistical technique for the dimension-wise decomposition of a square-integrable function $f: [0,1]^K \to \bb{R}$ into a sum of lower-dimensional functions $f_\mc{S}$ according to
\begin{align}\label{equ:dec1}
f(\bm{x}) = \sum\nolimits_{\mc{S} \subseteq \mc{K} } f_\mc{S}(\bm{x}_\mc{S}).
\end{align}
Here, each function $f_\mc{S}$ only depends on a subset of variables indexed by the set $\mc{S}\subseteq \mc{K}$ and the sum ranges over all $2^K$ subsets of $\mc{K}:=\{1,\hdots,K\}$.

A specific construction and algorithm was proposed in \cite{hooker2004discovering, KuSlWaWo10}, necessitating the computation of several multidimensional integrals. More precisely, the component functions are obtained by

\begin{align}\label{equ:anova_integral}
  f_\mc{S}(\bm{x}_\mc{S}) = \int\nolimits_{[0,1]^{K - \lvert \mc{S} \rvert}} f(\bm{x}) \ d \bm{x}_{ \mc{K} \backslash \mc{S}} - \sum\nolimits_{\mc{U} \subsetneq \mc{S} } f_\mc{U} (\bm{x}_\mc{U}),
\end{align}
where the first term represents an integral over a subset of variables and the second term subtracts all proper subsets in a manner similar to backfitting. Using this approach, one can demonstrate that all terms $f_\mc{S}$ are orthogonal with respect to the inner product $\langle f,g\rangle = \int f(\bm{x}) \cdot g(\bm{x}) \ d\bm{x}$ \cite{KuSlWaWo10}. Additionally, the construction also admits a decomposition of the functional variance
\begin{equation}\label{equ:full_variance}
    \sigma^2 = \int f^2(\bm{x}) d\bm{x} - \left( \int f d\bm{x} \right)^2
\end{equation}
into the sum of individual component variances
\begin{equation}\label{equ:component_variances}
    \sigma^2 = \sum\nolimits_{\mc{S} \subseteq \mc{K} } \sigma_\mc{S}^2 = \sum\nolimits_{\mc{S} \subseteq \mc{K} } \int f^2_\mc{S}(\bm{x}_\mc{S}) d\bm{x}_\mc{S}.
\end{equation}

Different approaches for calculating the ANOVA decomposition have been proposed in the literature, including methods based on random forests \cite{hutter2014} or Fourier basis functions \cite{Potts2021} which are analytically tractable. For more general function spaces, the numerical approximation of the integral \eqref{equ:anova_integral} is costly and can require at least a
number of $10^4$ function evaluations using a quasi-Monte Carlo
scheme in five dimensions \cite{owen2023practical}.

\subsection{Automatic Integration}
Analytical integration is considered more challenging than differentiation and closed-form solutions for general antiderivatives, i.e., indefinite integrals, are limited to a small class of functions, e.g., as obtained by the Risch algorithm \cite{risch1969problem}. 

Neural networks, being universal function approximators, can also be utilized for analytical integration within the framework of \emph{automatic integration}~\cite{Lindell2021} which sparked applications in neural radiance fields \cite{Gao2022}, pathloss prediction \cite{Limmer2023} and neural point processes \cite{Zhou2024}. This technique involves training the mixed partial derivative of a neural network so that integrals can be obtained by evaluating the trained network at the boundary points of the integration domain. We emphasize that the method \emph{enables the computation of $K$-dimensional integrals using $2^K$ evaluations of a neural network}, i.e., calculating an exact integral such as \eqref{equ:anova_integral} using only $32$ function evaluations in five dimensions.

\section{Neural-ANOVA Decomposition}
In this section, we present our main contribution, which provides a rapid and analytical evaluation of integrals over subspaces in the ANOVA decomposition (cf. \eqref{equ:anova_integral}), utilizing neural networks. \footnote{The implementation is available at \url{https://github.com/stli/neural-anova}.}
\subsection{Bivariate Example}
We begin by demonstrating the fundamental process of automatic integration using a sample bivariate function, \(f(x_1, x_2)\), to emphasize the differences in the training approach. Conventional neural network training  can be formulated as the following optimization problem
\begin{equation}
\min_{\bg{\theta}} \ \sum\nolimits_i \phi\Bigl( f(x_1^{(i)},x_2^{(i)}) - \text{NN}(\bg{\theta},x_1^{(i)},x_2^{(i)}) \Bigr),
\end{equation}
where \(\phi\) denotes an appropriate function to penalize the deviation between prediction and target such as the square or absolute value function.

In the proposed method, we aim to fit given samples of a function $f(x_1, x_2)$ while simultaneously calculating integrals over the input domain. The framework of automatic integration~\cite{Lindell2021} suggests training a neural network $\text{NN}(\bg{\theta}, x_1, x_2)$ by differentiating the network with respect to all its input coordinates, specifically evaluating its mixed partial derivative. The training process involves the optimization problem
\begin{equation*}\label{equ:nanova_training_0}
\min_{\bg{\theta}} \ \sum\nolimits_i \phi\Bigl( f(x_1^{(i)},x_2^{(i)}) - \frac{\partial^2}{\partial x_1 \partial x_2} \text{NN}(\bg{\theta},x_1^{(i)},x_2^{(i)}) \Bigr).
\end{equation*}
The term $\frac{\partial^2}{\partial x_1 \partial x_2} \text{NN}(\bg{\theta},x_1,x_2)$ can be compiled just-in-time and evaluated during the training process using standard techniques of automatic differentiation.

After successful optimization, the optimized neural network parameters, denoted as $\bg{\theta}^\star$, are obtained. Integrals can then be computed by evaluating the neural network at the corner points of the integration domain, $[l_1, u_1] \times [l_2, u_2]$ according to
\begin{align}
& \int_{l_1}^{u_1} \int_{l_2}^{u_2} f(x_1,x_2) \ dx_1 dx_2 \\
& = \text{NN}(\bg{\theta}^\star,l_1,l_2) - \text{NN}(\bg{\theta}^\star,u_1,l_2) \\
& \quad - \text{NN}(\bg{\theta}^\star,l_1,u_2) + \text{NN}(\bg{\theta}^\star,u_1,u_2) \nonumber \\
& := \text{NN}(\bg{\theta}^\star,x_1,x_2) \big\vert_{x_1,x_2 \in (l_1,u_1)\times (l_2,u_2) }.
\end{align}

\subsection{High-dimensional Generalization}
Next, we present the generalization of automatic integration to calculate higher-dimensional integrals. To this end, the training of a neural network $\text{NN}_{\bg{\theta}}(\bm{x}) : \bb{R}^K \to \bb{R}$ is formulated as the  optimization problem

\begin{equation}\label{equ:anova_train}
\min_{\bg{\theta}} \ \sum\nolimits_i \phi\Bigl( f(\bm{x}^{(i)}) - \frac{\partial^K}{\partial x_1 \cdots \partial x_K} \text{NN}_{\bg{\theta}}(\bm{x}^{(i)}) \Bigr).
\end{equation}

Then, we can establish the following relation between (i) the trained neural network, (ii) the general anti-derivative (integral) and (iii) the definite anti-derivative (integral) by using the fundamental theorem of calculus \cite{Mutze2004} according to 

\begin{gather} 
  f(\bm{x}) = \frac{\partial^K}{\partial x_1 \cdots \partial x_K} \text{NN}_{\bg{\theta}^\star}(\bm{x}) \Leftrightarrow \int f(\bm{x}) d\bm{x} = \text{NN}_{\bg{\theta}^\star}(\bm{x}) \nonumber \\
                                 \Leftrightarrow
    \int_{\bm{l}}^{\bm{u}} f(\bm{x}) d\bm{x} =  \sum_{\bm{x} \in (l_1,u_1)\times \cdots \times (l_K,u_K)} (-1)^s \text{NN}_{\bg{\theta}^\star}(\bm{x}),
\end{gather}

where $s$ denotes the number of occurrences of lower bounds in the evaluated expression.

Using this relation, we can verify that integration over a single variables (e.g. $x_1$) or a subset of variables (e.g. $x_2$,$x_3$) can be obtained for instance in the $3$-dimensional case by
\begin{align*}
&\int_{l_1}^{u_1} f(x_1, x_2, x_3) dx_1 = \frac{\partial^2}{\partial x_2 \partial x_3} \text{NN}(\bm{x}) \bigg\rvert_{x_1 \in (l_1,u_1) } \\
&\int f(x_1, x_2, x_3) dx_2 dx_3 = \frac{\partial}{\partial x_1} \text{NN}(\bm{x}) \bigg\rvert_{x_2,x_3 \in (l_2,u_2)\times (l_3,u_3) }.
\end{align*}

\subsection{Summary of Algorithm}
We now present the main result of this paper: a computational algorithm designed to train a neural network $\text{NN}_{\bg{\theta}^\star}(\bm{x})$, which allows for an analytical decomposition into lower-dimensional terms $\text{NN}_\mathcal{S}(\bm{x}_\mc{S})$. This method is termed \textit{Neural-ANOVA} and is summarized in Alg.~\ref{alg:neural-anova}.

In Alg. \ref{alg:neural-anova}, first the the network is trained via its mixed partial derivative according to the loss function \eqref{equ:anova_train}. Second, the integrals are calculated as the signed sum of mixed partial derivatives of the trained model according to
\begin{align}
    I_\mc{S}(\bm{x}_\mc{S}) &= \int\nolimits_{ \mc{S}^c} \text{NN}(\boldsymbol{x}) d\bm{x}_{\mc{S}^c} = \frac{\partial^{\lvert \mathcal{S} \rvert}}{\partial \boldsymbol{x}_\mathcal{S}} \text{NN}(\boldsymbol{x}) \bigg\vert_{\boldsymbol{x}_{\mc{S}^c} \in (0,1)^{\lvert \mc{S}^c \rvert}} \nonumber \\
    & =  \sum\nolimits_{\bm{x}_{\mc{S}^c} \in (0,1)^{\lvert \mc{S}^c \rvert}} (-1)^s \frac{\partial^{\lvert \mathcal{S} \rvert}}{\partial \boldsymbol{x}_\mathcal{S}} \text{NN}(\boldsymbol{x})  \label{equ:nanova_subspace_integral}.
\end{align}
Here, the sign exponent $s$ denotes the number of occurrences of lower bounds in the evaluated expression and $\mc{S}^c := \mc{K} \backslash \mc{S}$ is the complement of $\mc{S}$ over $\mc{K}$. In other words, the calculation of $I_\mc{S}(\bm{x}_\mc{S})$ entails differentiating w.r.t. the variables in the active-set $\mc{S}$ and evaluating at the $2^{\lvert \mc{S}^c \rvert}$ corner points of the inactive-set $\mc{S}^c$ so that the result is a function of only the active variables $\bm{x}_\mc{S}$. Lastly, the Neural-ANOVA component $\text{NN}_\mc{S}$ is obtained by the integral  \eqref{equ:nanova_subspace_integral} and subtracting of all proper subset components
\begin{align}\label{equ:nanova_subnetworks}
    \text{NN}_\mc{S}(\bm{x}_\mc{S}) = \int_{\mc{S}^c} \text{NN}(\boldsymbol{x}) d\bm{x}_{\mc{S}^c} - \sum_{\mc{U} \subsetneq \mc{S} } \text{NN}_\mc{U} (\bm{x}_\mc{U}).
\end{align}

The complete resulting algorithm to obtain \emph{Neural-ANOVA} is provided in Alg.~\ref{alg:neural-anova}. As can be seen from \eqref{equ:nanova_subspace_integral}, networks trained via its mixed partial derivative admit an exact analytical integration rule with complexity $\mathcal{O}(2^K)$ which compares favorably to inaccurate or expensive numerical approximation schemes. Moreover, only a single network needs to be trained to calculate integrals over arbitrary subspaces spanned by active variables $\bm{x}_\mc{S}$, $\mc{S} \subseteq \mc{K}$.

\begin{algorithm}
\setstretch{1.35}
\SetKwInput{KwData}{\textbf{Input}}
\SetKwInput{KwResult}{\textbf{Output}}
\KwData{Sampled function $f(\bm{x})\in \mc{L}_2([0,1]^K)$}
\KwResult{Nets $\{\text{NN}_\mc{S}\}_{\mc{S}\subseteq \mc{K}}$, variances $\{\sigma_\mc{S}\}^2_{\mc{S}\subseteq \mc{K}}$}
Obtain $\bg{\theta}^*$ by training $f(\bm{x}) \approx \frac{\partial^K}{\partial\bm{x}}\text{NN}_{\bg{\theta}}(\bm{x})$  \\
$\text{NN}_\emptyset := \sum_{\bm{x} \in (0,1)^K} (-1)^s \text{NN}_{\bg{\theta}}(\bm{x}) $ \\
$\sigma_\emptyset := 0$ \\
\For{$\mc{S}\subseteq \mc{K}$, $\mc{S}\neq \emptyset$}{
$\text{NN}_\mc{S}(\bm{x}_\mc{S}):= \frac{d^{\lvert \mathcal{S} \rvert}}{d \boldsymbol{x}_\mathcal{S}} \text{NN}(\boldsymbol{x}) \vert_{\boldsymbol{x}_{\mathcal{S}^c} \in (0,1)^{K - \lvert \mc{S} \rvert}} \newline
\qquad \qquad \quad - \sum_{\mc{U} \subsetneq \mc{S} } \text{NN}_\mc{U} (\bm{x}_\mc{U})$
$\sigma_\mc{S}^2 := \int_{[0,1]^{\lvert \mc{S} \rvert}}  \text{NN}^2_\mc{S}(\bm{x}_\mc{S}) \ d \bm{x}_\mc{S}$
}
$\sigma^2 := \int_{[0,1]^K} (\frac{d^K}{d\bm{x}} \text{NN})^2(\bm{x}) \ d\bm{x} - \text{NN}_\emptyset^2 \equiv \sum_{\mc{S} \subseteq\mc{K}} \sigma_\mc{S}^2$
\caption{Neural-ANOVA decomposition of $f$, adapted from \cite{KuSlWaWo10}.}
\label{alg:neural-anova}
\end{algorithm}

One approach to calculate the mixed partial derivative in \eqref{equ:anova_train} supported by standard automatic differentiation frameworks is to apply nested differentiation. This approach is used in this work as we observe satisfactory runtime and numerical stability up to moderate dimension $K \leq 10$. While the implementation of this approach is straight forward e.g. in the automatic differentiation framework JAX~\cite{jax2018github}, it requires traversing the original computation graph multiple times which results in redundant computations as is noted in \cite{Hoffmann2016,Bettencourt2019}. 
We highlight that (approximate) calculation of higher-order partial derivatives via the Taylor series is an active field of research \cite{Bettencourt2019,shi2024stochastic} to further improve the scalability of training and evaluating mixed partial derivatives.

\subsection{Numerical Example}
This section presents a concise numerical example for a common test-function from sensitivity analysis, namely the $3$-dimensional \emph{Ishigami}-function
\begin{align}
f(\bm{x})=\sin(x_1)+a \sin^2(x_2) + b x_3^4\sin(x_1),
\end{align}
with $a=7, b=0.1$. 

We normalize the input and output domain and present the generated data for $x_3=0$ in Fig. \ref{fig:anova-teaser}. The optimization problem to obtain the trained network $\text{NN}(\bm{x}):=\text{NN}_{\bg{\theta}^\star}(\bm{x})$ is defined as
\begin{align}
\min_{\bg{\theta}} \ \sum_i \Bigl( f(x_1,x_2,x_3) - \frac{\partial ^3}{\partial x_1 \partial x_2 \partial x_3} \text{NN}(x_1,x_2,x_3) \Bigr)^2.
\end{align}
Next, we find the terms of the Neural-ANOVA decomposition using the trained network according to
\begin{align*}
\text{NN}_\emptyset &= \text{NN}(u_1,u_2,u_3) - \text{NN}_{\bg{\theta}^\star}(u_1,l_2,u_3) \\ \nonumber
&-\text{NN}(l_1,u_2,u_3) + \text{NN}(l_1,l_2,u_3) -\text{NN}(u_1,u_2,l_3) \\ \nonumber
&+ \text{NN}(u_1,l_2,l_3) + \text{NN}(l_1,u_2,l_3) - \text{NN}(l_1,l_2,l_3) \\
\text{NN}_1(x_1) &= \frac{\partial}{\partial x_1} \text{NN}(x_1,x_2,x_3)\bigg\rvert_{x_2,x_3 \in (l_2,u_2)\times (l_3,u_3) } - \text{NN}_\emptyset \\
\text{NN}_{1,2}&(x_1,x_2) = \frac{\partial}{\partial x_1\partial x_2} \text{NN}(x_1,x_2,x_3)\bigg\rvert_{x_3 \in (l_3,u_3)} \nonumber \\
&- \text{NN}_\emptyset - \text{NN}_1(x_1) - \text{NN}_2(x_2)
\end{align*}

\begin{table}
\centering
\small
    \begin{tabular}{ccccccc}
    \toprule
        $\mathcal{S}$ & $\{1\}$ & $\{2\}$ & $\{3\}$ & $\{1,2\}$ & $\{1,3\}$ & $\{2,3\}$ \\
        \midrule
        Ref. & 3.14E-1 & 4.42E-1 & 0.0 & 0.0 & 2.44E-1 & 0.0  \\
        Est. & 3.05E-1 & 4.39E-1 & 6E-9 & 1E-10 & 2.56E-1 & 8E-9 \\
        \bottomrule
    \end{tabular}
    \caption{True sensitivities and numerical estimates calculated using Neural-ANOVA for the Ishigami function.}
    \label{tab:ishigami_sensitivities}
\end{table}

\begin{figure}
\centering \subfigure[]{
  \includegraphics[width=0.45\linewidth]{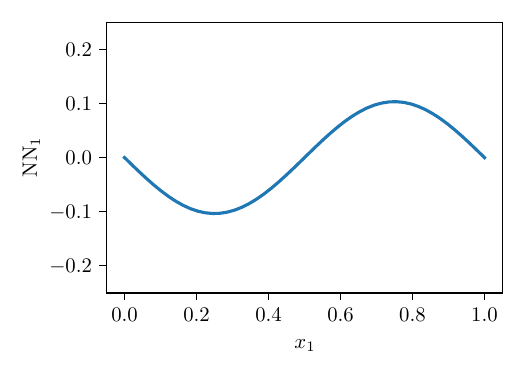}
} \subfigure[]{
  \includegraphics[width=0.45\linewidth]{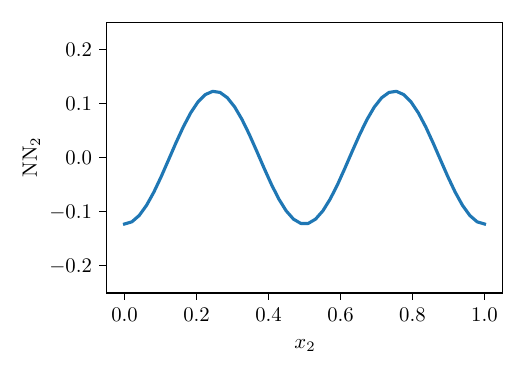}
}
\centering \subfigure[]{
  \includegraphics[width=0.45\linewidth]{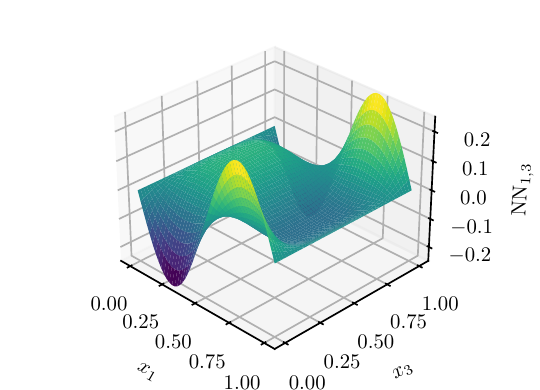}
} \subfigure[]{
  \includegraphics[width=0.45\linewidth]{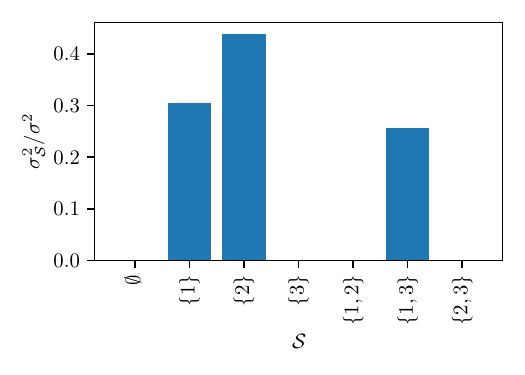}
}
\caption{(a-c) Plots of $\text{NN}_\mathcal{S}(\bm{x}_\mathcal{S})$ for $\mc{S}= \{1\},\{2\},\{1,3\}$, and (d) sensitivities $\sigma_\mathcal{S}$ for the Ishigami function.}
\label{fig:ishigami_functions}
\end{figure}

Finally, we can evaluate and illustrate the decomposed function (cf. Fig. \ref{fig:ishigami_functions}) according to Alg.~\ref{alg:neural-anova} and may further calculate component variances $\sigma_\mc{S}$, e.g., using a Monte Carlo estimate. We see in Tab.~\ref{tab:ishigami_sensitivities} that the component variances match well with their known closed form expressions  \cite{sobol1999use}.

\begin{table}
\centering
\small
    \begin{tabular}{ccccccc}
    \toprule
         & ISH & CIR & PST & ASN & CCP & CCS \\
     \midrule
        features & 3 & 6 & 7 & 5 & 4 & 8 \\
        samples & 10000 & 10000 & 10000 & 1503 & 9568 & 1030 \\
    \bottomrule
    \end{tabular}
    \caption{Test function and dataset overview.}
    \label{tab:dataset_overview}
\end{table}

\begin{table*}[t]
\centering
\small
\begin{tabular}{l || l | lllll | l}
\toprule
 & MLP & $\text{N-ANOVA}_{K}$ & $\text{N-ANOVA}_{4}$ & $\text{N-ANOVA}_{3}$ & $\text{N-ANOVA}_{2}$ & $\text{N-ANOVA}_{1}$ & NAM \\
\midrule
ISH & \begin{tabular}{@{}c@{}} 1.7E-04 \\\footnotesize{$\pm$0.5E-04}\end{tabular} & \begin{tabular}{@{}c@{}} 1.2E-04 \\\footnotesize{$\pm$0.2E-04}\end{tabular} & \begin{tabular}{@{}c@{}} 1.2E-04 \\\footnotesize{$\pm$0.2E-04}\end{tabular} & \begin{tabular}{@{}c@{}} 1.2E-04 \\\footnotesize{$\pm$0.2E-04}\end{tabular} & \begin{tabular}{@{}c@{}} {1.2E-04} \\\footnotesize{$\pm$0.2E-04}\end{tabular} & \begin{tabular}{@{}c@{}} {5.06E-02} \\\footnotesize{$\pm$0.04E-02}\end{tabular} & \begin{tabular}{@{}c@{}} {5.08E-02} \\\footnotesize{$\pm$0.05E-02}\end{tabular} \\ \\
CIR & \begin{tabular}{@{}c@{}} 5.8E-05 \\\footnotesize{$\pm$1.5E-05}\end{tabular} & \begin{tabular}{@{}c@{}} 1.3E-04 \\\footnotesize{$\pm$0.3E-04}\end{tabular} & \begin{tabular}{@{}c@{}} 1.1E-04 \\\footnotesize{$\pm$0.2E-04}\end{tabular} & \begin{tabular}{@{}c@{}} 1.0E-04 \\\footnotesize{$\pm$0.2E-04}\end{tabular} & \begin{tabular}{@{}c@{}} {1.1E-04} \\\footnotesize{$\pm$0.2E-04}\end{tabular} & \begin{tabular}{@{}c@{}} {1.59E-02} \\\footnotesize{$\pm$0.01E-02}\end{tabular} & \begin{tabular}{@{}c@{}} {1.61E-02} \\\footnotesize{$\pm$0.02E-02}\end{tabular} \\ \\
PST & \begin{tabular}{@{}c@{}} 5.2E-05 \\\footnotesize{$\pm$0.8E-05}\end{tabular} & \begin{tabular}{@{}c@{}} 1.65E-04 \\\footnotesize{$\pm$0.2E-04}\end{tabular} & \begin{tabular}{@{}c@{}} 2.52E-04 \\\footnotesize{$\pm$0.07E-04}\end{tabular} & \begin{tabular}{@{}c@{}} 2.96E-03 \\\footnotesize{$\pm$0.05E-03}\end{tabular} & \begin{tabular}{@{}c@{}} 1.62E-02 \\\footnotesize{$\pm$0.03E-02}\end{tabular} & \begin{tabular}{@{}c@{}} {3.94E-02} \\\footnotesize{$\pm$0.04E-02}\end{tabular} & \begin{tabular}{@{}c@{}} {3.86E-02} \\\footnotesize{$\pm$0.06E-02}\end{tabular} \\ \\
ASN & \begin{tabular}{@{}c@{}} 4.4E-02 \\\footnotesize{$\pm$0.2E-02}\end{tabular} & \begin{tabular}{@{}c@{}} 9.0E-02 \\\footnotesize{$\pm$0.3E-02}\end{tabular} & \begin{tabular}{@{}c@{}} 9.0E-02 \\\footnotesize{$\pm$0.3E-02}\end{tabular} & \begin{tabular}{@{}c@{}} 1.00E-01 \\\footnotesize{$\pm$0.09E-01}\end{tabular} & \begin{tabular}{@{}c@{}} {1.19E-01} \\\footnotesize{$\pm$0.07E-01}\end{tabular} & \begin{tabular}{@{}c@{}} 1.67E-01 \\\footnotesize{$\pm$0.07E-01}\end{tabular} & \begin{tabular}{@{}c@{}} {1.23E-01} \\\footnotesize{$\pm$0.02E-01}\end{tabular} \\ \\
CCP & \begin{tabular}{@{}c@{}} 5.33E-02 \\\footnotesize{$\pm$0.08E-02}\end{tabular} & \begin{tabular}{@{}c@{}} 5.73E-02 \\\footnotesize{$\pm$0.05E-02}\end{tabular} & \begin{tabular}{@{}c@{}} 5.73E-02 \\\footnotesize{$\pm$0.05E-02}\end{tabular} & \begin{tabular}{@{}c@{}} {5.74E-02} \\\footnotesize{$\pm$0.05E-02}\end{tabular} & \begin{tabular}{@{}c@{}} {5.77E-02} \\\footnotesize{$\pm$0.05E-02}\end{tabular} & \begin{tabular}{@{}c@{}} 5.95E-02 \\\footnotesize{$\pm$0.06E-02}\end{tabular} & \begin{tabular}{@{}c@{}} {5.68E-02} \\\footnotesize{$\pm$0.06E-02}\end{tabular} \\ \\
CCS & \begin{tabular}{@{}c@{}} 7.4E-02 \\\footnotesize{$\pm$0.2E-02}\end{tabular} & \begin{tabular}{@{}c@{}} 1.03E-01 \\\footnotesize{$\pm$0.06E-01}\end{tabular} & \begin{tabular}{@{}c@{}} 1.03E-01 \\\footnotesize{$\pm$0.06E-01}\end{tabular} & \begin{tabular}{@{}c@{}} 1.04E-01 \\\footnotesize{$\pm$0.06E-01}\end{tabular} & \begin{tabular}{@{}c@{}} 1.06E-01 \\\footnotesize{$\pm$0.06E-01}\end{tabular} & \begin{tabular}{@{}c@{}} 1.51E-01 \\\footnotesize{$\pm$0.2E-01}\end{tabular} & \begin{tabular}{@{}c@{}} {7.1E-02} \\\footnotesize{$\pm$0.2E-02}\end{tabular} \\
\bottomrule
\end{tabular}
\caption{
    \label{tab:performance_comparison}
Performance comparison of various models: $\text{N-ANOVA}_{d}$ (user-defined interactions of order $d \in \{1,2,3,\hdots,K\}$ \eqref{equ:dec1}), Neural Additive Model ($\text{NAM}$, no interactions) and Multi-Layer Perceptron (MLP, full order interactions). \text{N-ANOVA} is trained as a single network per dataset and may include user defined interactions. Performance is measured using Root Mean Squared Error (RMSE) on the holdout set, with lower RMSE values indicating better performance.
}
\end{table*}
\begin{figure*}[t]
\centering \subfigure[]{
  \includegraphics[width=0.234\linewidth]{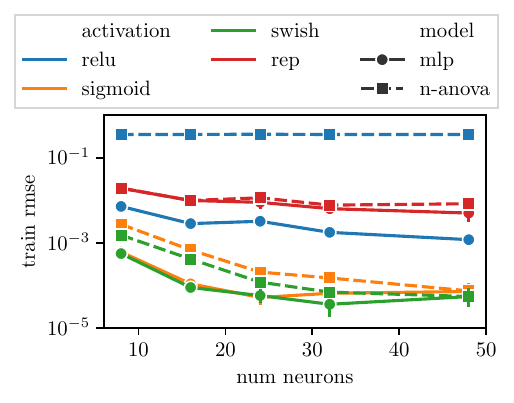}
} \subfigure[]{
  \includegraphics[width=0.234\linewidth]{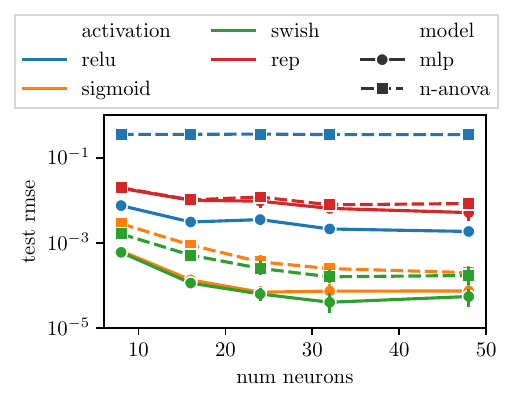}
}
\subfigure[]{
  \includegraphics[width=0.234\linewidth]{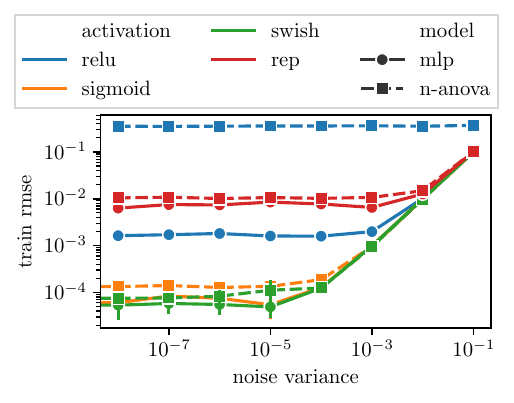}
} \subfigure[]{
  \includegraphics[width=0.234\linewidth]{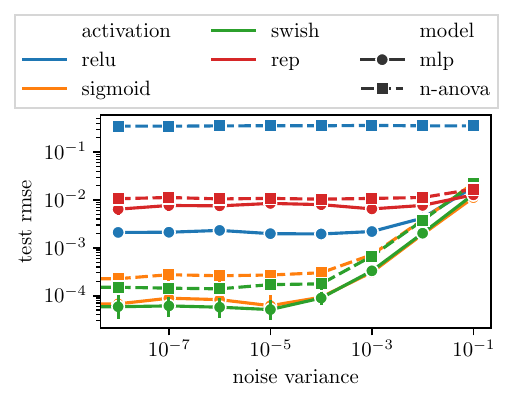}
}
\caption{Ablation on (a)-(b) training and testing error for varying number of hidden layer neurons and (c)-(d) for varying level of additive training noise and varying activation functions on the Piston dataset.}
\label{fig:ablation_piston_noise}
\end{figure*}

\section{Experiments}
This section presents the results of numerical experiments performed on test functions and datasets. We use \emph{Ishigami} (ISH), \emph{OTL Circuit} (CIR) and \emph{Piston} (PST) from the \emph{UQTestFuns}-library \cite{wicaksono2023uqtestfuns}.

As benchmark datasets we use Airfoil Self-Noise (ASN), Combined Cycle Power Plant (CCP) and Concrete Compressive Strength (CCS) datasets \cite{asuncion2007uci}. We provide an overview of the considered datasets in Tab.~\ref{tab:dataset_overview}. In all cases the data is split into $60 / 20 / 20$ ratio for training, validation, and, testing, respectively.

We use JAX \cite{jax2018github} to implement both Neural-ANOVA and an MLP baseline with the following standard architectures: (i) $3$ layers with $32$ neurons and \textrm{sigmoid} activation, and ablations with (ii)  $\{8,16,32,48\}$ hidden neurons and $\{\textrm{sigmoid}, \textrm{relu}, \textrm{swish}, \textrm{rep}\}$ activation where $\textrm{rep}$ denotes the rectified polynomial function.
The default architecture is used for ISH, CIR and PST where we observe a moderate increase of training time of $121.3\,$s for MLP compared to $857.6\,$s for Neural-ANOVA on the 7-dimensional PST dataset, respectively. For the ASN, CCP, and CCS datasets, we observe the necessity of regularization to improve generalization. Our empirical findings indicate that a two-layer architecture with \textrm{rep} activation, $16$ neurons and \(\ell_2\)-weight regularization provides satisfactory results for ASN and CCP, but led to a small number of divergent runs, which were excluded from the analysis. We also report results for Neural Additive Models (NAMs), which consist of three layers with 32 neurons and \textrm{relu} activation, following the JAX implementation\footnote{\url{https://github.com/Habush/nam_jax}}.


In Tab.~\ref{tab:performance_comparison}, we report the root MSE (RMSE) and standard error on the test set, based on 10 runs with different random seeds and validation early stopping with adam and bfgs optimizers. We find that MLP and $\text{N-ANOVA}_K$ (all interactions) perform comparable in the large data regime on the test functions ISH, CIR, PST.
For datasets with a small sample count, NAMs demonstrate slightly superior generalization to univariate \(\text{N-ANOVA}_1\). This performance can be matched by \(\text{N-ANOVA}_{2}\) where bivariate interactions are included. However, \(\text{N-ANOVA}\) shows performance deterioration for small sample sizes in higher dimensions, specifically for the CCS dataset with sparse data in the largest dimension \(K=8\) which suggests further research on architectures with improved generalization given by its mixed partial derivative.

Finally, Fig.~\ref{fig:ablation_piston_noise} shows an ablation study on different activation functions and additive noise levels in order to verify approximation properties and numerical stability. The results indicate that mixed partial derivatives slightly reduce expressive power compared to the same non-differentiated architecture, but also admit an improved approximation with increasing layer size and an overall comparable robustness to noise. Notably, mixed partial derivatives of a \textrm{relu} architecture admit a significant loss in expressive power which we conjecture results from vanishing differentials, while the \textrm{rep} activation shows promising robustness to higher noise levels of \(10^{-1}\).
%
\section{Conclusion}
{
\widowpenalty=10000000
\clubpenalty=10000000
In this paper, we present an efficient method for computing the functional ANOVA decomposition using neural networks. We derive a novel learning problem by fitting the mixed partial derivative to the training data to compute integrals over subspaces appearing in the ANOVA decomposition. Our approach is empirically validated on various test functions from uncertainty quantification and benchmark datasets, confirming the accuracy of the functional decomposition. We also show that Neural-ANOVA  can include a variable order of interaction effects and achieves comparable performance to standard approaches such as neural additive models. Furthermore, the method provides a mathematically well-founded way to analyze interaction effects of trained or distilled models, and provides deeper insights into the expressive power and noise robustness of neural architectures under mixed differentiation. Further research may address scalability improvements of training and inference, e.g., by integrating recently proposed methods for exact or stochastic Taylor-mode derivatives \cite{Bettencourt2019,shi2024stochastic} or an additional outer function related to nomographic approximation \cite{limmer2015simple}. The implementation of Neural-ANOVA used to generate the results is made available with the paper.
}

\bibliographystyle{IEEEbib}
\bibliography{root}

\end{document}